\documentclass[10pt,twocolumn,letterpaper]{article}

\usepackage[]{wacv}
\usepackage{amsmath,amssymb,amsfonts}
\usepackage{graphicx}
\usepackage{booktabs}
\usepackage{multirow}
\usepackage{makecell}
\usepackage{array}
\usepackage{tabularx}
\usepackage{pifont}
\usepackage{subcaption}
\usepackage{url}
\usepackage{xspace}
\usepackage{microtype}
\usepackage{placeins}

\newcommand{\method}{AViS-Mamba\xspace}
\newcommand{\R}{\mathbb{R}}

\newcommand{\aav}{\mathcal{L}_{\mathrm{AAV}}}

\definecolor{wacvblue}{rgb}{0.21,0.49,0.74}
\usepackage[pagebackref,breaklinks,colorlinks,allcolors=wacvblue]{hyperref}

\def\wacvPaperID{225}
\def\confName{WACV}
\def\confYear{2027}

\title{\method: Adaptive Visual Steering of Audio State-Space Dynamics for Violence Detection}

\author{Damith Chamalke Senadeera$^{1,2}$, Dimitrios Kollias$^{1,2}$, Gregory Slabaugh$^{1,2}$\\
$^{1}$Queen Mary University of London, UK\\
$^{2}$Digital Environment Research Institute (DERI), London, UK\\
{\tt\small \{d.c.senadeera, d.kollias, g.slabaugh\}@qmul.ac.uk}
}

\begin{document}
\maketitle

\begin{abstract}

Automatic violence detection from video is challenging because violent interactions may be distant, occluded, or only partially visible. Audio can provide complementary evidence for violent events that are difficult to recognize from visual information alone. However, audio itself may be absent, dubbed, or dominated by environmental noise, making the central challenge not whether to incorporate audio but how to adapt reliance on it according to the visual scene. We introduce \emph{AViS-Mamba}, an audiovisual Mamba-based architecture in which the visual stream directly governs the behavior of the audio stream. At each layer of the audio encoder, a compact visual representation produces a modulation vector that conditions the encoder's internal temporal operators together with a routing gate that regulates the strength of this visual intervention. Rather than fusing or reweighting features after they have been extracted, visual context directly shapes the temporal dynamics of the audio encoder. We further propose Adaptive AV-InfoNCE, a contrastive objective that learns to balance the audio-to-video and video-to-audio alignment directions rather than weighting them uniformly. On the audio-valid NTU-CCTV and DVD benchmarks, AViS-Mamba establishes state-of-the-art results, attaining 88.59\% and 75.74\% accuracy. We demonstrate that adaptive visual conditioning consistently outperforms fixed routing and improves performance under degraded and missing-audio conditions. Layer-wise analysis further reveals that the model adapts the audio stream selectively across network depth rather than applying a single global routing policy.

\end{abstract}

\begin{figure}[t]
    \centering
    \includegraphics[width=0.80\columnwidth]{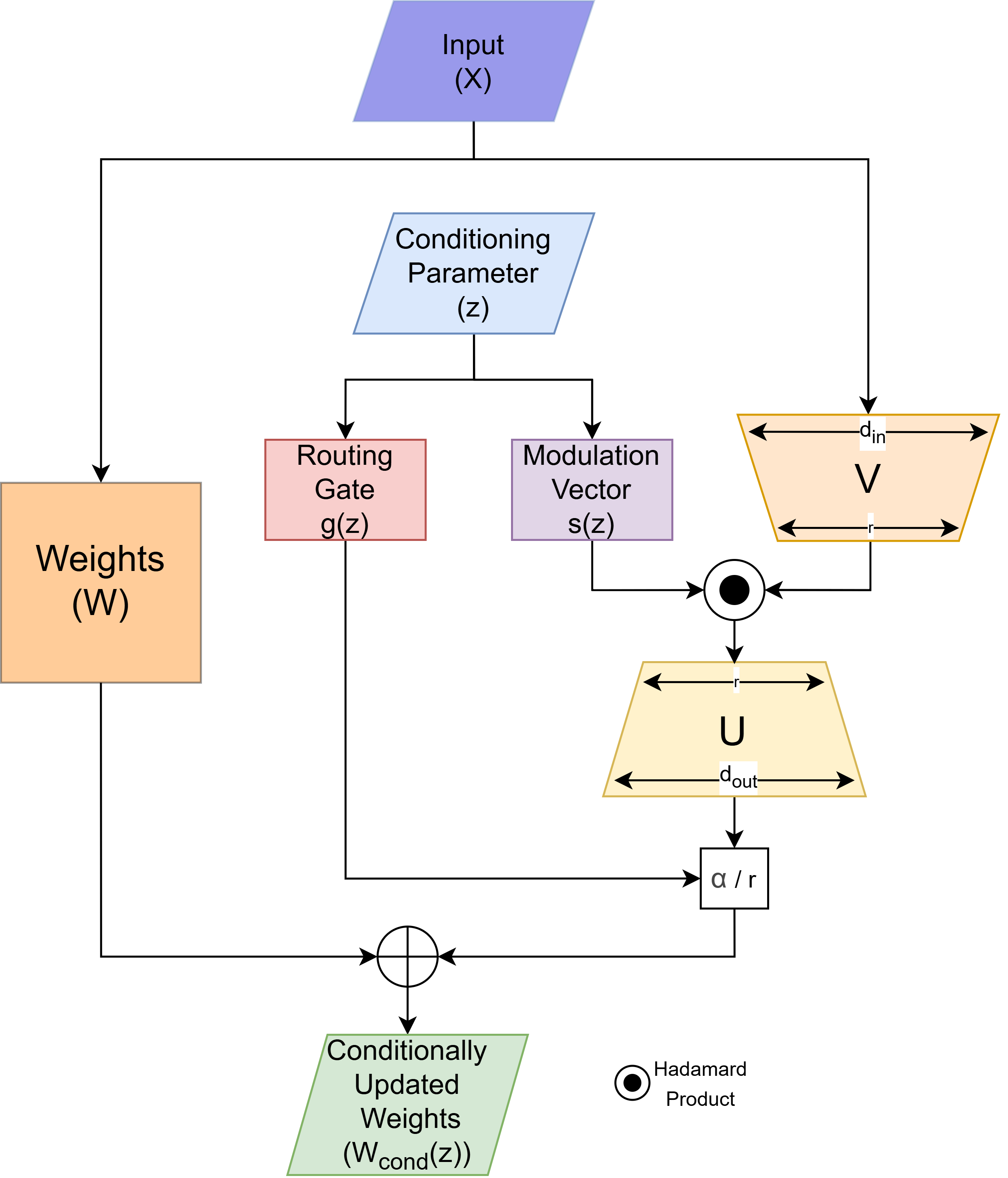}
    \caption{Visual-conditioned operator modulation. A clip-level controller produces a routing gate and a rank-channel modulation vector; together they form a sample-dependent correction to the jointly trained base operator. The low-rank basis bounds cost.}
    \label{fig:mechanism}
\end{figure}

\section{Introduction}
Violence recognition is difficult when decisive actions are small, occluded, blurred, or embedded in crowded scenes. Audio can resolve this ambiguity through impacts, screams, gunshots, breaking objects, and crowd reactions. Yet the same soundscape may be absent, dubbed, dominated by traffic or wind, or only weakly synchronized with the visible event. The central challenge is therefore not simply to add audio to the violence detection approach, but to determine how visual evidence should control the way audio is processed.

Most audiovisual systems combine representations after feature extraction through concatenation, affine modulation, bottleneck attention, or cross-attention \cite{avslowfast2020,mbt2021,lavish2023,tim2024,cavmaesync2025}. We ask a different question: \emph{can one modality directly reshape the temporal operators used by another?} In a selective state-space model (SSM), input-dependent parameters determine what enters the state, how quickly it evolves, and what is read out. Conditioning these generators changes the dynamics before the recurrent scan, rather than reweighting features after the transformation has already occurred.

\begin{figure*}[t]
    \centering
    \includegraphics[width=\textwidth]{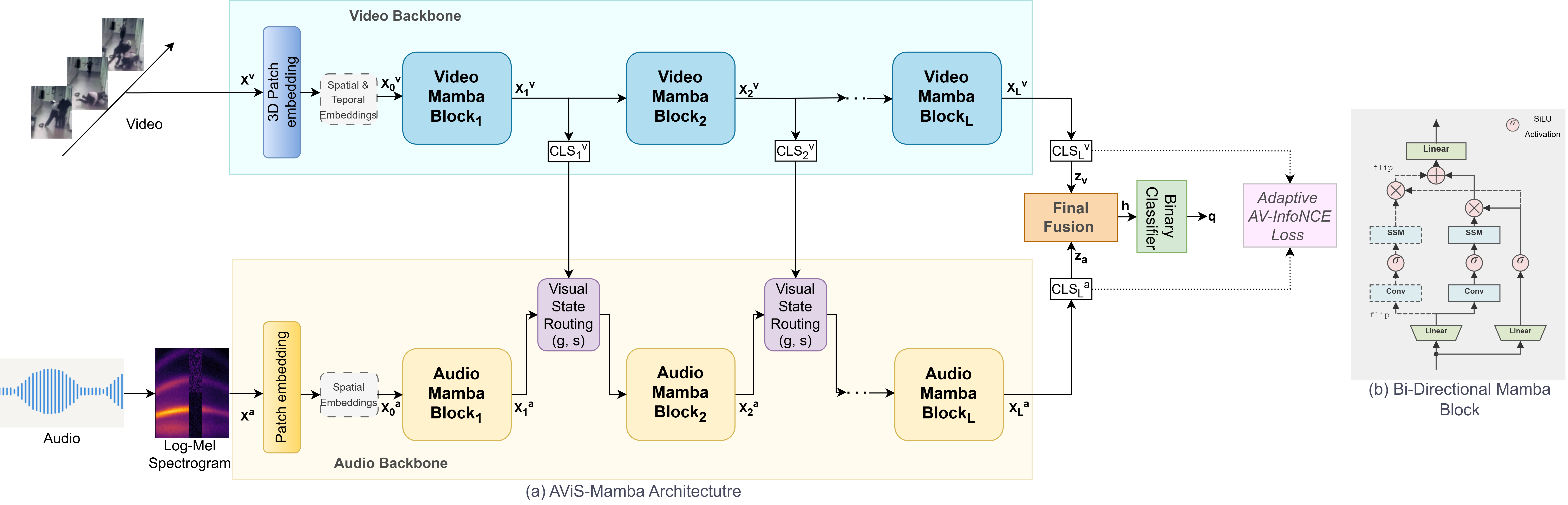}
    \caption{\method overview. \textbf{(a)}~Full architecture: The video backbone processes the video input $\mathbf{X}^v$, while audio backbone processes the log-mel spectrogram of $\mathbf{X}^a$ in parallel. At each layer $\mathrm{CLS}_\ell^v$ from the video backbone modulates the audio backbone via each layer specific visual state routing block using the routing gate $g$ and the modulation vector $s$. The final video and audio descriptors ($\mathbf{z}_v$, $\mathbf{z}_a$) are concatenated for classification and regularized by Adaptive AV-InfoNCE.
    \textbf{(b)}~Bidirectional Mamba block shared by both backbones.}
    \label{fig:architecture}
\end{figure*}

We introduce \method, shown in Fig.~\ref{fig:architecture}, in which video acts as the semantic anchor and audio as a complementary stream. At every depth, the visual CLS token generates a channel-wise modulation vector and a routing gate for two AudioMamba operators, namely the joint generator of the selective state parameters and the dedicated step-size generator, yielding a sample-dependent audio operator at every layer without matching audio and video token rates or constructing dense cross-modal attention. The low-rank factorization serves only as an efficient basis for generating this dynamic correction rather than as a parameter-efficient fine-tuning adapter, since all backbones, controllers, and classifiers are optimized jointly and the update changes with the visual content of every clip. We therefore formulate the contribution as \emph{visual-conditioned state-space operator routing}. Alongside this, we introduce Adaptive AV-InfoNCE, which replaces the equal average of audio-to-video and video-to-audio alignment losses with a bounded learned scalar that adapts their balance while keeping both directions active, giving the model a controlled mechanism for learning directional cross-modal alignment.

Our experiments establish both that the method works and why. Across five seeds, \method sets a new state-of-the-art on the audio-valid NTU-CCTV and DVD benchmarks. To understand the contribution of audio and of learning the routing weight, we run a controlled intervention that fixes the routing gate $g$ to constant values at test time; we evaluate every clip in the unfiltered DVD split as a robustness stress test; we measure zero-shot cross-dataset transfer; and we analyze the learned routing gate across depth. The detailed intervention tables and the full layer-wise routing gate analysis are provided in the supplement, with the key findings summarized here.

Our contributions are:
\begin{itemize}
    \item We introduce visual-conditioned operator routing inside the selective state-space parameter generators of an audio encoder, including direct control of the step-size pathway.
    \item We propose Adaptive AV-InfoNCE, a bounded learned balance of the two directional audio-video contrastive objectives.
    \item We achieve state-of-the-art results on NTU-CCTV and DVD datasets and, through controlled routing intervention, unfiltered-data, transfer, and layer-wise routing gate analysis, quantify both the contribution and the behavior of adaptive routing.
\end{itemize}

\section{Related Work}

\textbf{Violence detection.}
Fully supervised violence recognition has progressed from handcrafted motion descriptors and 3D CNNs to Transformer and SSM video encoders \cite{hockeyfight,review_1,videoswintrf,uniformerv2,cuenet2024,fusemambavd2025}. A complementary line studies weakly supervised localization on long videos. XD-Violence introduced large-scale audiovisual weak supervision \cite{hlnet2020}, followed by modality-aware contrastive learning \cite{macilsd2022}, bottleneck fusion \cite{msbt2024}, and alignment-first multimodal methods \cite{mavd2025}.
However, these methods either fuse audio late as an auxiliary signal or rely on weak video-level labels that obscure fine-grained audiovisual interactions. Our setting is clip-level binary recognition from temporal annotations: rather than targeting temporal localization, it exploits this finer granularity to learn tightly coupled audiovisual dynamics that are hard to be captured from coarser weakly supervised pipelines.

\textbf{Audiovisual fusion.}
SlowFast-AV, MBT, LAVISH, TIM, and CAV-MAE Sync represent strong feature-, token-, and pretraining-based fusion paradigms \cite{avslowfast2020,mbt2021,lavish2023,tim2024,cavmaesync2025}. Violence-specific alternatives model audiovisual dependency through attention or retain performance through cross-modal causal distillation \cite{avdatt2023,mfvd2024}. Recent work also studies reconstruction-based audiovisual discrepancy modeling \cite{auvire2026} and domain-generalizable multimodal action recognition \cite{ji2026alignment}.
These approaches fuse at the feature or token level, leaving each modality's internal temporal operators unchanged and treating the two streams as broadly symmetric. In contrast, we condition the audio stream's internal temporal operators and preserve a directional video-to-audio hierarchy, embedding cross-modal influence inside the sequence dynamics rather than appending it as an external fusion stage.

\textbf{State-space and conditional modulation.}
Mamba makes SSM parameters input-dependent while retaining linear sequence complexity \cite{mamba2024}. VideoMamba and AudioMamba adapt this principle to visual and acoustic inputs \cite{videomamba2024,audiomamba2024}; other recent work uses Mamba for cross-view sensor fusion \cite{millimamba2026}. FiLM modulates features through input-conditioned affine transformations \cite{perez2018film}, while LoRA uses a static low-rank weight correction \cite{lora2022}.
Our update instead modulates the low-rank factors by another modality on every sample and inserts them into the generators of selective SSM dynamics, uniting conditional modulation and low-rank adaptation so the video stream directly reshapes how the audio SSM selects and propagates information.

\section{Method}
\subsection{Selective state-space preliminaries}
For a sequence $\mathbf{x}_{1:T}$, a selective SSM~\cite{mamba2024} updates a latent state and emits an output as
\begin{equation}
\mathbf{h}_t=\bar{\mathbf{A}}_t\mathbf{h}_{t-1}+\bar{\mathbf{B}}_t\mathbf{x}_t,
\qquad
\mathbf{y}_t=\mathbf{C}_t\mathbf{h}_t+\mathbf{D}\mathbf{x}_t ,
\end{equation}
where $\mathbf{h}_t \in \mathbb{R}^{N}$ is the hidden state, $\bar{\mathbf{A}}_t$ is the discretized state-transition matrix, $\bar{\mathbf{B}}_t$ admits the input into the state, $\mathbf{C}_t$ reads the state out, and $\mathbf{D}$ is a skip connection. The discretized matrices depend on an input-conditioned step size $\boldsymbol{\Delta}_t$ and continuous-time parameters. In AudioMamba, an internal projection jointly produces the token-dependent raw step-size term $\mathbf{d}^{\mathrm{raw}}_t$ together with $\mathbf{B}_t, \mathbf{C}_t$ from the current internal input feature $\tilde{\mathbf{x}}_t$,
\begin{equation}
[\mathbf{d}^{\mathrm{raw}}_t,\mathbf{B}_t,\mathbf{C}_t]
=\mathbf{W}_{x}\tilde{\mathbf{x}}_t,
\end{equation}
where $\mathbf{W}_{x}$ is the joint input linear projection. A dedicated projection $\mathbf{W}_{\Delta}$ with bias $\mathbf{b}_{\Delta}$ then maps $\mathbf{d}^{\mathrm{raw}}_t$ to a positive step size,
\begin{equation}
\boldsymbol{\Delta}_t=
\operatorname{softplus}(\mathbf{W}_{\Delta}\mathbf{d}^{\mathrm{raw}}_t+\mathbf{b}_{\Delta}).
\end{equation}
We keep the selective scan unchanged and condition the joint input projection and the step-size projection.

\subsection{Method initialization}

Our framework couples two state-space backbones through a directional, video-conditioned mechanism. The video stream is encoded by VideoMamba \cite{videomamba2024}: for sample $i$, the clip $\mathbf{X}_i^v\in\mathbb{R}^{3\times T\times H\times W}$ is patchified, prepended with a CLS token, augmented with positional embeddings $\mathbf{p}_s,\mathbf{p}_t$, and processed into the layer-$\ell$ sequence $\mathbf{X}_i^{v,\ell}=[\mathbf{x}_{i,\mathrm{cls}}^{v,\ell},\mathbf{x}_{i,1}^{v,\ell},\dots,\mathbf{x}_{i,N_v}^{v,\ell}]$. From each layer we take the normalized CLS token $\mathbf{c}_{v,i}^{\ell}=\mathrm{LN}(\mathbf{x}_{i,\mathrm{cls}}^{v,\ell})$ as a layer-wise guidance signal, treating video as the anchor modality. In parallel, the audio waveform $\mathbf{X}_i^a$ is converted to a log-mel spectrogram, patchified with a CLS token and positional embeddings, and processed by AudioMamba \cite{audiomamba2024}.

\subsection{Visual-conditioned operator modulation}
Let $\mathbf{c}^{\ell}_{v,i}$ be the normalized VideoMamba CLS token for clip $i$ at layer $\ell$, which serves as the cross-modal conditioning signal. For a base linear operator $\mathbf{W}^{\ell}$, the visual controller predicts a channel-wise modulation vector $\mathbf{s}^{\ell}_i$ and a scalar routing gate $g^{\ell}_i$,
\begin{equation}
\mathbf{s}^{\ell}_i=\tanh(\operatorname{MLP}^{\ell}(\mathbf{c}^{\ell}_{v,i})),
\quad
g^{\ell}_i=\sigma((\mathbf{w}^{\ell}_g)^\top\mathbf{c}^{\ell}_{v,i}+b^{\ell}_g),
\end{equation}
where $\operatorname{MLP}^{\ell}$ maps the CLS token to the modulation vector, and $\mathbf{w}^{\ell}_g$ and $b^{\ell}_g$ are the weight vector and bias of the gate. The controller applies
\begin{equation}
\mathcal{F}^{\ell}(\mathbf{x};\mathbf{c}_{v,i}^{\ell})
=
\mathbf{W}^{\ell}\mathbf{x}
+
g_i^{\ell}\frac{\alpha}{r}\mathbf{U}^{\ell}
\left[
(\mathbf{V}^{\ell}\mathbf{x})\odot\mathbf{s}_i^{\ell}
\right],
\label{eq:dynamic_operator}
\end{equation}
where $\mathbf{U}^{\ell}\in\R^{d_{\mathrm{out}}\times r}$ and $\mathbf{V}^{\ell}\in\R^{r\times d_{\mathrm{in}}}$ are the up- and down-projection factors, $\alpha$ is a scaling factor, and $r$ is the rank. Here $r$ bounds the cost, $\mathbf{s}^{\ell}_i$ selects rank channels, and $g_i^{\ell}$ scales the update. Equation~\ref{eq:dynamic_operator} therefore defines a different operator for every clip and layer. The base weights and the routing branch are optimized jointly; the factorization is used to make conditional operator generation tractable, not as a parameter-efficient fine-tuning procedure. For an operator of size $d_{\mathrm{out}}\times d_{\mathrm{in}}$, the dynamic correction costs $\mathcal{O}(r(d_{\mathrm{in}}+d_{\mathrm{out}}))$ per token rather than materializing a dense clip-specific matrix.

We instantiate separate controllers for the joint input projection and the step-size projection:
\begin{align}
\mathbf{p}^{a,\ell}_{i,t}
&=\mathbf{W}^{\ell}_{x}\tilde{\mathbf{x}}^{a,\ell}_{i,t}
+g^{\ell}_{x,i}\frac{\alpha_x}{r_x}\mathbf{U}^{\ell}_{x}
\left[
(\mathbf{V}^{\ell}_{x}\tilde{\mathbf{x}}^{a,\ell}_{i,t})
\odot\mathbf{s}^{\ell}_{x,i}
\right], \\
\boldsymbol{\delta}^{\mathrm{pre},\ell}_{i,t}
&=\mathbf{W}^{\ell}_{\Delta}\mathbf{d}^{\mathrm{raw},\ell}_{i,t}
+g^{\ell}_{\Delta,i}\frac{\alpha_\Delta}{r_\Delta}\mathbf{U}^{\ell}_{\Delta}
\left[
(\mathbf{V}^{\ell}_{\Delta}\mathbf{d}^{\mathrm{raw},\ell}_{i,t})
\odot\mathbf{s}^{\ell}_{\Delta,i}
\right] \nonumber \\
&\quad +\mathbf{b}^{\ell}_{\Delta}.
\end{align}
where $\tilde{\mathbf{x}}^{a,\ell}_{i,t}$ is the internal audio input token feature at token index $t$, $\mathbf{p}^{a,\ell}_{i,t}$ is the resulting joint projection output $[\mathbf{d}^{\mathrm{raw},\ell}_{i,t},\mathbf{B}^{\ell}_{i,t},\mathbf{C}^{\ell}_{i,t}]$, and $\boldsymbol{\delta}^{\mathrm{pre},\ell}_{i,t}$ is the step-size pre-activation. The subscripts $x$ and $\Delta$ denote the input-projection and step-size controllers respectively, each with its own up- and down-projection factors $(\mathbf{U}, \mathbf{V})$, scaling $\alpha$, rank $r$, modulation vector $\mathbf{s}$, and routing gate $g$.
Because the same visual CLS token is broadcast to all audio tokens within a layer, we obtain scene-aware but stable modulation without explicit token-level cross-attention or modality-rate matching. Modulating both input projection and step-size projection allows the visual stream to shape not only $\mathbf{B}_t$ and $\mathbf{C}_t$ but also the rate at which the audio state evolves.

\textbf{A new fusion interface.}
The novelty lies in \emph{where} cross-modal control is applied. Unlike feature fusion, cross-attention, or a static low-rank update that applies the same correction to every clip, Eq.~\ref{eq:dynamic_operator} uses the visual state to generate a clip-specific, layer-wise modulation inserted directly into the generators of the audio stream's selective state parameters. A low-rank basis keeps this operator economical.

\subsection{Fusion and Adaptive AV-InfoNCE}
After the final layer, we use the last visual CLS token $\mathbf{z}_{v,i}=\mathbf{c}_{v,i}^{L}$ and the last audio CLS token $\mathbf{z}_{a,i}=\mathbf{c}_{a,i}^{L}$ as clip-level descriptors. They are concatenated and fed to a binary classifier,
\begin{equation}
\mathbf{h}_i=[\mathbf{z}_{v,i};\mathbf{z}_{a,i}], \qquad
q_i=\mathbf{W}_{\mathrm{cls}}\mathbf{h}_i+b_{\mathrm{cls}},
\end{equation}
where $q_i$ is the violence logit. Training combines binary classification with an adaptive audio-video alignment. The classification loss $\mathcal{L}_{\mathrm{cls}}$ is the standard binary cross-entropy loss with final logits. For Adaptive AV-InfoNCE, we project the final audio and video embeddings into a shared space:
\begin{equation}
\mathbf{z}_i^a=f_a(\mathbf{z}_{a,i})\in \mathbb{R}^{D}, \qquad
\mathbf{z}_i^v=f_v(\mathbf{z}_{v,i})\in \mathbb{R}^{D},
\end{equation}
normalize them:
\begin{equation}
\tilde{\mathbf{z}}_i^a=\frac{\mathbf{z}_i^a}{\lVert \mathbf{z}_i^a\rVert_2}, \qquad
\tilde{\mathbf{z}}_i^v=\frac{\mathbf{z}_i^v}{\lVert \mathbf{z}_i^v\rVert_2},
\end{equation}
and compute a contrastive loss with learned temperature $\tau=\exp(\rho)$:
\begin{equation}
\mathcal{L}_{a\rightarrow v}=
-\frac{1}{B}\sum_{i=1}^{B}
\log\frac{\exp(\tilde{\mathbf{z}}_i^a\!\cdot\!\tilde{\mathbf{z}}_i^v/\tau)}{\sum_{j=1}^{B}\exp(\tilde{\mathbf{z}}_i^a\!\cdot\!\tilde{\mathbf{z}}_j^v/\tau)},
\end{equation}
\begin{equation}
\mathcal{L}_{v\rightarrow a}=
-\frac{1}{B}\sum_{i=1}^{B}
\log\frac{\exp(\tilde{\mathbf{z}}_i^v\!\cdot\!\tilde{\mathbf{z}}_i^a/\tau)}
{\sum_{j=1}^{B}\exp(\tilde{\mathbf{z}}_i^v\!\cdot\!\tilde{\mathbf{z}}_j^a/\tau)},
\end{equation}
where $B$ is the batch size. When combining $\mathcal{L}_{a\rightarrow v}$ \& $\mathcal{L}_{v\rightarrow a}$, instead of taking a fixed average, we learn a bounded directional weight
\begin{equation}
\omega=m+(1-2m)\sigma(\beta), \qquad
\aav=\omega\mathcal{L}_{a\rightarrow v}+(1-\omega)\mathcal{L}_{v\rightarrow a},
\label{eq:aav}
\end{equation}
where $m$ is the minimum per-direction weight, $\beta$ is a learnable gate logit, and $\sigma$ is the sigmoid. We use $m=0.05$ and initialize $\omega=0.5$, so training begins from balanced InfoNCE while each direction always retains at least $5$\% weight. The bounded parameterization prevents the optimizer from eliminating the currently harder direction, and the gate logit $\beta$ is excluded from weight decay. We log $\omega$, both directional losses, and the learned logit scale, but do not interpret $\omega$ as modality importance. The complete objective is
\begin{equation}
\mathcal{L}=\mathcal{L}_{\mathrm{BCE}}+\lambda\aav,
\end{equation}
where $\mathcal{L}_{\mathrm{BCE}}$ is the binary cross-entropy classification loss and $\lambda$ balances classification against cross-modal alignment.

\section{Experimental Setup}
\paragraph{Datasets and audiovisual (A/V) protocol.}
We use two temporally annotated violence-detection datasets: NTU-CCTV~\cite{ntucctv2019} and DVD~\cite{dvd2025}. NTU-CCTV contains CCTV and mobile-camera videos, while DVD contains more diverse real-world and media footage. For both datasets, we convert maximal contiguous violent and non-violent intervals into clips.

Our primary benchmark uses an audio-valid protocol. This is necessary because the A/V SOTA methods that we compare our method with, assume that both modalities are present and scene-related. Evaluating clips with missing, silent, dubbed, narrated, or off-scene audio would introduce a separate missing-modality problem, where different masking or imputation choices could dominate the comparison. We therefore apply a prediction-blind, model-independent filter before training and evaluation: clips are removed if they lack an audio stream, have peak audio level below $-80$ dB, or contain predominantly dubbed, narrated, or off-scene sound. This protocol isolates A/V learning on genuinely paired audio--video evidence and is not based on clip difficulty or model outputs.

To prevent leakage, all splits are performed at the source-video level. After filtering, NTU-CCTV retains 4,715 of 5,823 raw clips, and DVD retains 2,245 of 2,648 raw clips. The final train/test partitions are reported in Table~\ref{tab:splits}. In addition to the audio-valid benchmark, we evaluate the full unfiltered DVD test split of 661 clips as a native robustness stress test. We will release the split manifests and per-clip filtering decisions.

\begin{table}[t]
\centering
\caption{Clip counts.}
\label{tab:splits}
\small
\setlength{\tabcolsep}{4pt}
\begin{tabular}{llrrr}
\toprule
Dataset & Split & Violent & Non-violent & Total \\
\midrule
\multirow{2}{*}{NTU-CCTV} & Train & 1502 & 2026 & 3528 \\
 & Test & 507 & 680 & 1187 \\
\midrule
\multirow{3}{*}{DVD} & Train & 698 & 965 & 1663 \\
 & Test (filtered) & 260 & 322 & 582 \\
 & Test (unfiltered) & 279 & 382 & 661 \\
\bottomrule
\end{tabular}
\end{table}

\begin{table*}[t]
\centering
\caption{Performance comparison between AViS-Mamba and SOTA methods on the audio-valid splits. Results are reported as mean $\pm$ std over 5 seeds and are statistically significant ($p<<0.001$). Best and second-best values are highlighted in bold and underlined, respectively.}
\label{tab:main_results}
\scriptsize
\setlength{\tabcolsep}{2.6pt}
\renewcommand{\arraystretch}{1.08}
\resizebox{\textwidth}{!}{%
\begin{tabular}{lllcccccccc}
\toprule
\multirow{2}{*}{Model} & \multirow{2}{*}{Arch.} & \multirow{2}{*}{Mod.} &
\multicolumn{4}{c}{NTU-CCTV (\%)} & \multicolumn{4}{c}{DVD (\%)} \\
\cmidrule(lr){4-7}\cmidrule(lr){8-11}
& & & Acc. & F1-V & F1-NV & Macro-F1 & Acc. & F1-V & F1-NV & Macro-F1 \\
\midrule
ResNet-50 & CNN & A & 69.17$\pm$0.42 & 63.03$\pm$0.58 & 73.56$\pm$0.32 & 68.29$\pm$0.45 & 58.76$\pm$0.54 & 54.71$\pm$0.81 & 62.14$\pm$0.63 & 58.43$\pm$0.55 \\
AST & Transf. & A & 71.02$\pm$0.25 & 65.39$\pm$0.25 & 75.07$\pm$0.23 & 70.23$\pm$0.24 & 60.82$\pm$0.24 & 56.82$\pm$0.55 & 64.15$\pm$0.17 & 60.48$\pm$0.28 \\
AudioMamba & SSM & A & 72.82$\pm$0.30 & 68.58$\pm$0.42 & 76.06$\pm$0.23 & 72.32$\pm$0.32 & 62.61$\pm$0.46 & 58.75$\pm$0.62 & 65.81$\pm$0.40 & 62.28$\pm$0.48 \\
\midrule
SlowFast & CNN & V & 76.41$\pm$0.48 & 72.76$\pm$0.55 & 79.20$\pm$0.43 & 75.98$\pm$0.49 & 63.57$\pm$0.38 & 59.54$\pm$0.40 & 66.87$\pm$0.38 & 63.21$\pm$0.38 \\
VideoSwin-B & Transf. & V & 78.77$\pm$0.21 & 75.49$\pm$0.29 & 81.28$\pm$0.19 & 78.38$\pm$0.22 & 65.81$\pm$0.32 & 62.66$\pm$0.42 & 68.46$\pm$0.27 & 65.56$\pm$0.33 \\
UniFormer-V2 & Hybrid & V & 80.37$\pm$0.32 & 77.18$\pm$0.41 & 82.78$\pm$0.26 & 79.98$\pm$0.33 & 67.35$\pm$0.32 & 64.42$\pm$0.42 & 69.84$\pm$0.27 & 67.13$\pm$0.33 \\
CUE-Net & Hybrid & V & 82.06$\pm$0.38 & 79.18$\pm$0.42 & 84.23$\pm$0.34 & 81.71$\pm$0.38 & 70.03$\pm$0.51 & 67.61$\pm$0.66 & 72.12$\pm$0.43 & 69.86$\pm$0.52 \\
VideoMamba-M & SSM & V & 82.22$\pm$0.27 & 79.00$\pm$0.34 & 84.59$\pm$0.24 & 81.80$\pm$0.27 & 69.42$\pm$0.44 & 64.25$\pm$0.65 & 73.27$\pm$0.41 & 68.76$\pm$0.46 \\
VideoMambaPro-M & SSM & V & 80.96$\pm$0.34 & 77.44$\pm$0.42 & 83.53$\pm$0.28 & 80.49$\pm$0.35 & 68.90$\pm$0.32 & 63.87$\pm$0.45 & 72.70$\pm$0.26 & 68.29$\pm$0.34 \\
DBVideoMamba & SSM & V & 83.99$\pm$0.19 & 81.45$\pm$0.21 & 85.93$\pm$0.17 & 83.69$\pm$0.19 & 71.82$\pm$0.32 & 70.07$\pm$0.40 & 73.38$\pm$0.28 & 71.72$\pm$0.33 \\
\midrule
SlowFast-AV & CNN & A/V & 79.09$\pm$0.32 & 75.86$\pm$0.42 & 81.56$\pm$0.25 & 78.71$\pm$0.33 & 65.05$\pm$0.45 & 61.69$\pm$0.70 & 67.87$\pm$0.35 & 64.78$\pm$0.48 \\
TIM & Transf. & A/V & 83.03$\pm$0.22 & 80.20$\pm$0.24 & 85.16$\pm$0.20 & 82.68$\pm$0.22 & 68.83$\pm$0.38 & 65.97$\pm$0.42 & 71.25$\pm$0.35 & 68.61$\pm$0.38 \\
MBT & Transf. & A/V & 85.22$\pm$0.26 & \underline{82.79$\pm$0.34} & 87.06$\pm$0.20 & 84.92$\pm$0.27 & 73.33$\pm$0.22 & \underline{71.43$\pm$0.24} & 75.00$\pm$0.22 & 73.21$\pm$0.22 \\
LAVISH & Transf. & A/V & \underline{86.42$\pm$0.41} & 82.23$\pm$0.57 & \underline{89.01$\pm$0.33} & \underline{85.62$\pm$0.45} & \underline{74.33$\pm$0.52} & 69.07$\pm$0.66 & \underline{78.06$\pm$0.43} &  \underline{73.57$\pm$0.55} \\
CAV-MAE Sync & Transf. & A/V & 84.89$\pm$0.19 & 82.08$\pm$0.23 & 86.93$\pm$0.17 & 84.51$\pm$0.20 & 73.85$\pm$0.22 & 70.84$\pm$0.25 & 76.28$\pm$0.21 & 73.56$\pm$0.23 \\
DepMamba & SSM & A/V & 78.15$\pm$0.30 & 74.64$\pm$0.41 & 80.80$\pm$0.23 & 77.72$\pm$0.32 & 67.11$\pm$0.38 & 63.26$\pm$0.43 & 70.23$\pm$0.35 & 66.75$\pm$0.38 \\
\midrule
\textbf{\method} & SSM & A/V & \textbf{88.59$\pm$0.23} & \textbf{86.23$\pm$0.33} & \textbf{90.26$\pm$0.17} & \textbf{88.25$\pm$0.25} & \textbf{75.74$\pm$0.33} & \textbf{72.76$\pm$0.11} & \textbf{78.12$\pm$0.52} & \textbf{75.44$\pm$0.27} \\
\bottomrule
\end{tabular}}
\end{table*}

\textbf{Implementation.}
VideoMamba-Middle is initialized from the public Kinetics-400 checkpoint and receives 64 RGB frames at $224{\times}224$. AudioMamba receives 16\,kHz log-mel spectrograms with 128 mel bins, a 40\,ms window, and a 10\,ms hop. We use AdamW, learning rate $10^{-4}$, weight decay 0.05, 5 warm-up epochs, and 55 total epochs. The final setting uses $r=8$, $\alpha=2$, $\lambda=0.4$, and a 128-unit modulation MLP. Evaluation uses 4 temporal segments and 3 spatial crops. Full preprocessing, augmentation, and optimizer settings are in the supplementary.

\textbf{Evaluation Metrics.}
We report accuracy, violent-class F1 (F1-V), non-violent F1 (F1-NV), and Macro-F1. Unless otherwise stated, benchmark results are reported as mean $\pm$ sample standard deviation over five independent seeds.

For the main benchmark, we also test the Macro-F1 difference between \method{} and the strongest baseline using a two-tailed paired $t$-test over the five random seeds.

\section{Results}
\subsection{State-of-the-art Comparison}
We first evaluate whether AViS-Mamba improves clip-level violence detection under the audio-valid protocol, where audio and visual streams are present and scene-related. Table~\ref{tab:main_results} compares AViS-Mamba to audio-only, video-only and A/V state-of-the-art (SOTA) on NTU-CCTV and DVD.
AViS-Mamba achieves the strongest performance among all evaluated methods (accuracy, Macro-F1, F1-V and F1-NV) on both datasets. On NTU-CCTV, it reaches 88.59\% accuracy and 88.25\% Macro-F1, surpassing the best SOTA, LAVISH (A/V method), by 2.17 points in accuracy and 2.63 points in Macro-F1.
On DVD, AViS-Mamba reaches 75.74\% accuracy and 75.44\% Macro-F1, surpassing LAVISH by 1.41 points in accuracy and 1.88 points in Macro-F1.
Our method significantly outperforms the second-best baseline (LAVISH) in Macro-F1 under a two-tailed paired $t$-test across five random seeds, on both NTU-CCTV ($+2.63$, $t(4) = 18.88$, $p < 10^{-4}$) and DVD ($+1.88$, $t(4) = 10.65$, $p < 10^{-3}$).

The improvements are also accompanied by strong seed stability. On both NTU-CCTV and DVD, AViS-Mamba has lower standard deviation than LAVISH across all reported metrics. This lower variability suggests that the proposed operator-routing mechanism improves average performance while also producing more consistent results across independent runs.

\subsection{Performance--Computational Cost trade-off}

We next examine whether the performance gain of \method is achieved at a reasonable computational cost. Table~\ref{tab:efficiency} compares \method with SOTA A/V methods on DVD in terms of Macro-F1, parameter count and FLOPs. At the mean level, \method improves over LAVISH by $1.88$ Macro-F1 points while using 80M fewer parameters and 360 fewer GFLOPs. These results indicate that the proposed operator-routing mechanism improves performance without relying on a larger or more expensive model.

\begin{table}[h]
\centering
\caption{
DVD performance--cost comparison. Macro-F1 is reported as mean $\pm$ std over 5 seeds.}
\label{tab:efficiency}

\setlength{\tabcolsep}{2.4pt}
\begin{tabular}{lrrrr}
\toprule
Model & Params (M) & GFLOPs & Macro-F1. \\
\midrule
SlowFast-AV & \textbf{38.5} & \textbf{680} & 64.78$\pm$0.48 \\
TIM & 93.7 & 1520 & 68.61$\pm$0.38  \\
MBT & 174.0 & 1860 & 73.21$\pm$0.22 \\
LAVISH & 238.0 & 2170 & \underline{73.57$\pm$0.55} \\
CAV-MAE Sync & 179.2 & 1950 & 73.56$\pm$0.23 \\
DepMamba & \underline{49.6} & \underline{830} & 66.75$\pm$0.38 \\
\textbf{\method} & 158.0 & 1810 & \textbf{75.44$\pm$0.27} \\
\bottomrule
\end{tabular}
\end{table}

\subsection{Ablation Studies}
We next isolate which design choices are responsible for the performance of \method. Table~\ref{tab:connection_types} first compares different cross-modal coupling mechanisms and directions on DVD, while keeping the backbone, training schedule and evaluation protocol fixed. This analysis asks whether the proposed visual-conditioned operator is more effective than standard feature- or token-level fusion, and whether video should control audio, audio should control video, or both streams should be coupled symmetrically. The results show a clear directional pattern. Across all tested mechanisms, Video$\rightarrow$Audio coupling consistently outperforms Audio$\rightarrow$Video and criss-cross coupling. The proposed visual-conditioned operator achieves the best result. This suggests that, for violence detection, visual context provides the more reliable control signal, while audio is most useful as a complementary stream whose temporal dynamics are adapted according to the visual scene.

\begin{table}[h]
\centering
\caption{
Layer-wise coupling mechanisms on DVD (Acc in \%). ``Static LR'' is a content-independent low-rank update, whereas the proposed operator uses sample-conditioned visual modulation.}
\label{tab:connection_types}
\small
\setlength{\tabcolsep}{3pt}
\resizebox{\columnwidth}{!}{%
\begin{tabular}{lrrr}
\toprule
Mechanism & V$\rightarrow$A & A$\rightarrow$V & Criss-cross \\
\midrule
Cross-attention & 70.44 & 61.17 & 60.65 \\
Concatenation & 70.10 & 60.82 & 61.51 \\
FiLM & 70.79 & 60.14 & 60.31 \\
Static LR update & 70.96 & 60.65 & 61.00 \\
Visual-conditioned operator & \textbf{71.48} & \textbf{61.17} & \textbf{61.86} \\
\bottomrule
\end{tabular}}
\end{table}

Table~\ref{tab:configuration_path} follows the selected Video$\rightarrow$Audio design via a controlled configuration path. Starting from the same operator with routing disabled, enabling the route increases DVD accuracy. Increasing the update scale to $\alpha=2$ further improves accuracy, showing that sufficient conditional update capacity is important.
Finally, adding Adaptive AV-InfoNCE raises performance further (from $74.23$ to $75.74$), showing that directional audio--visual alignment provides an additional gain once the routed operator is in place.

\begin{table}[h]
\centering
\caption{Stepwise configuration path for the selected Video$\rightarrow$Audio design on DVD.}
\label{tab:configuration_path}
\small
\setlength{\tabcolsep}{3.2pt}
\resizebox{\columnwidth}{!}{%
\begin{tabular}{lcccl r}
\toprule
Direction & Route & $r$ & $\alpha$ & Objective & Acc. \\
\midrule
V$\rightarrow$A & off & 8 & 1 & BCE & 69.76 \\
V$\rightarrow$A & on & 8 & 1 & BCE & 71.48 \\
V$\rightarrow$A & on & 8 & 2 & BCE & 74.23 \\
V$\rightarrow$A & on & 8 & 2 & Adaptive ($\lambda=0.4$) & \textbf{75.74$\pm$0.33} \\
\bottomrule
\end{tabular}}
\end{table}

\begin{table*}[h]
\centering
\caption{Representative DVD prediction flips. All clips contain violent events. In the upper rows, audio corrects visual-only false negatives; in the lower rows, misleading audio turns visual-only correct predictions into false negatives.}
\label{tab:qualitative_flip}
\scriptsize
\setlength{\tabcolsep}{2pt}
\renewcommand{\arraystretch}{1.04}
\begin{tabular}{p{6.7cm}p{3.05cm}p{3.05cm}cc}
\toprule
\textbf{Frames} & \textbf{Visual evidence} & \textbf{Audio evidence} & \textbf{V-only} & \textbf{\method} \\
\midrule
\multicolumn{5}{l}{\textit{Audio helps}} \\
\raisebox{-0.48\height}{%
  \includegraphics[width=1.90cm]{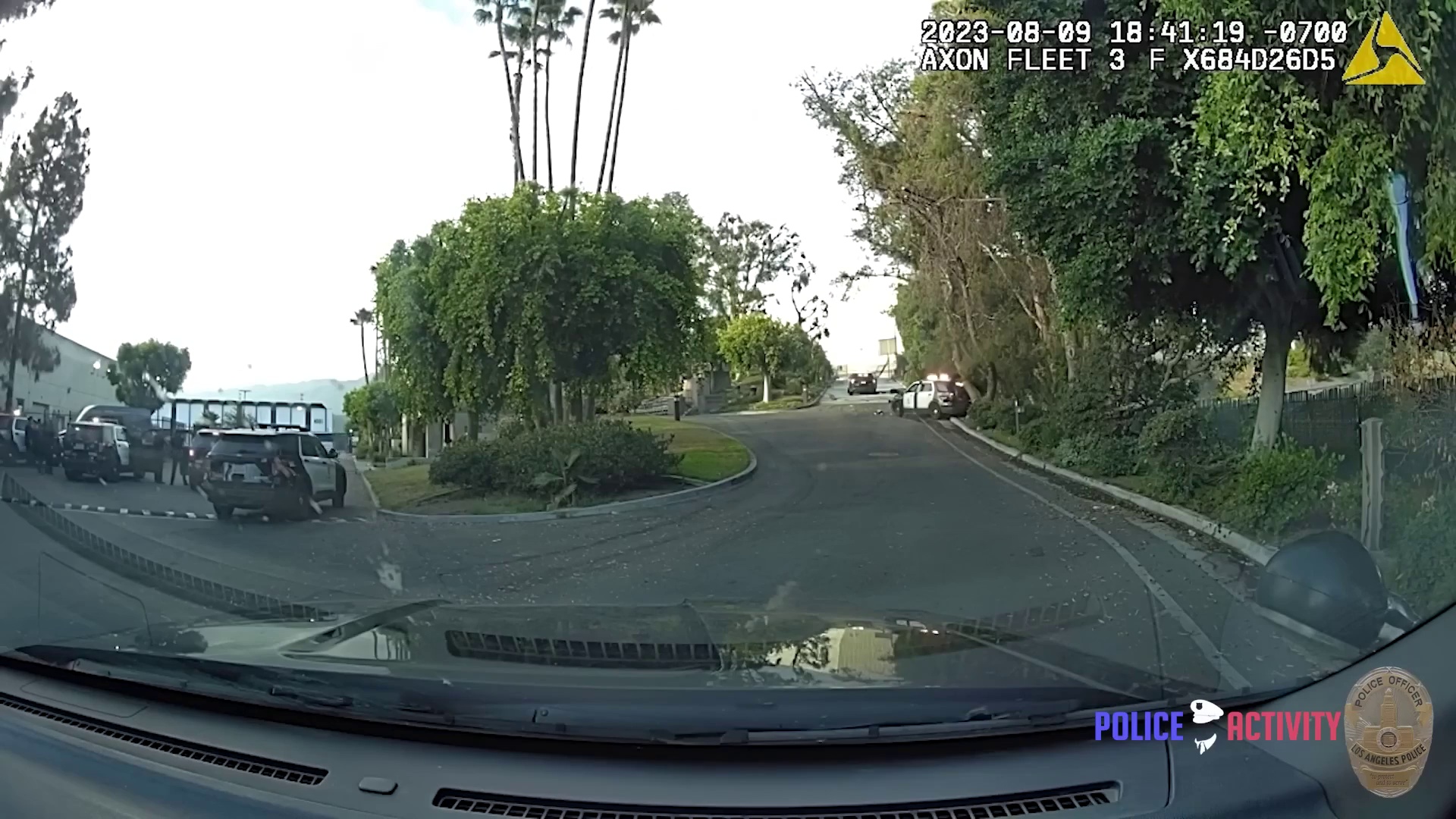}\hspace{1pt}%
  \includegraphics[width=1.90cm]{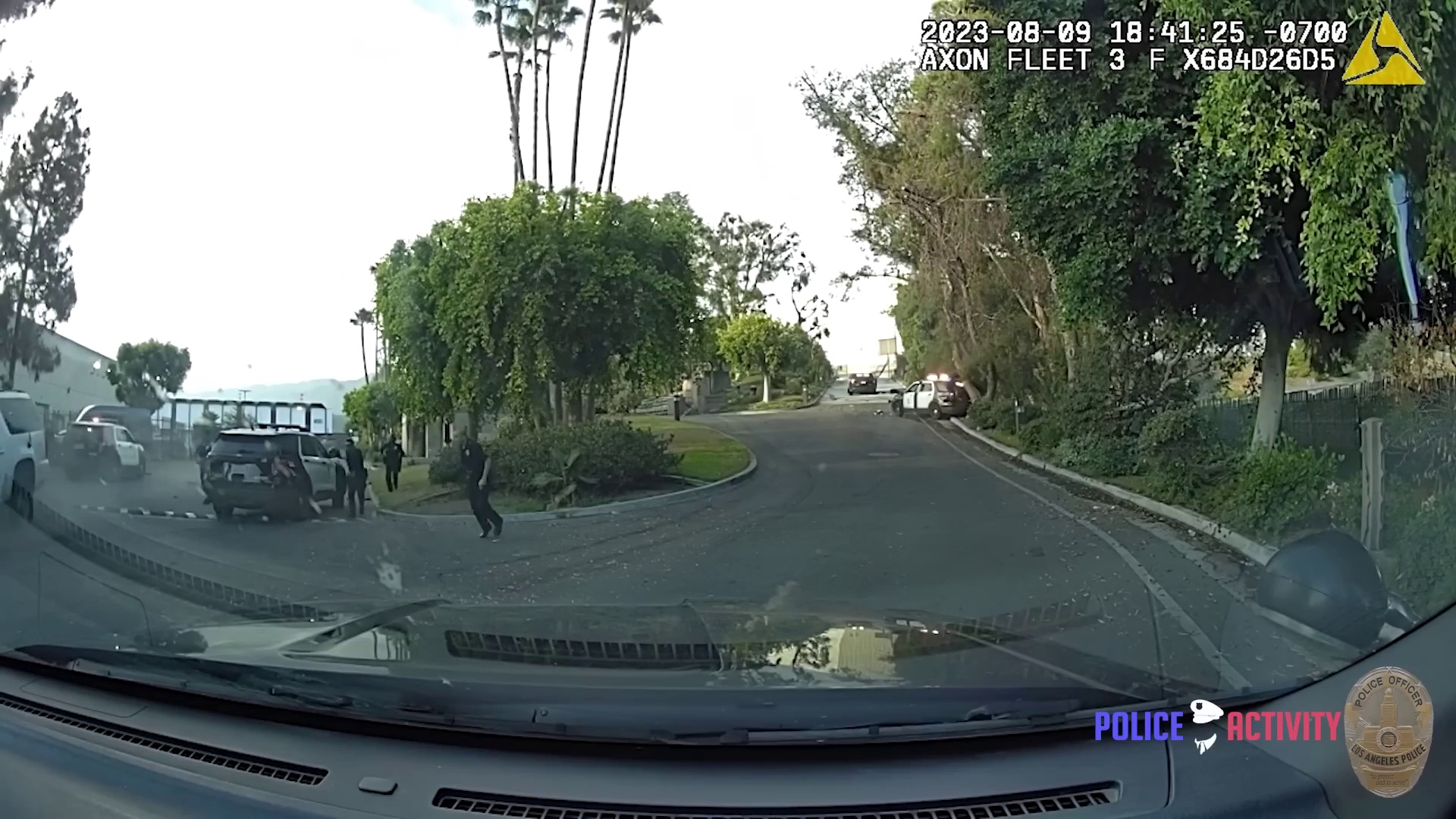}\hspace{1pt}%
  \includegraphics[width=1.90cm]{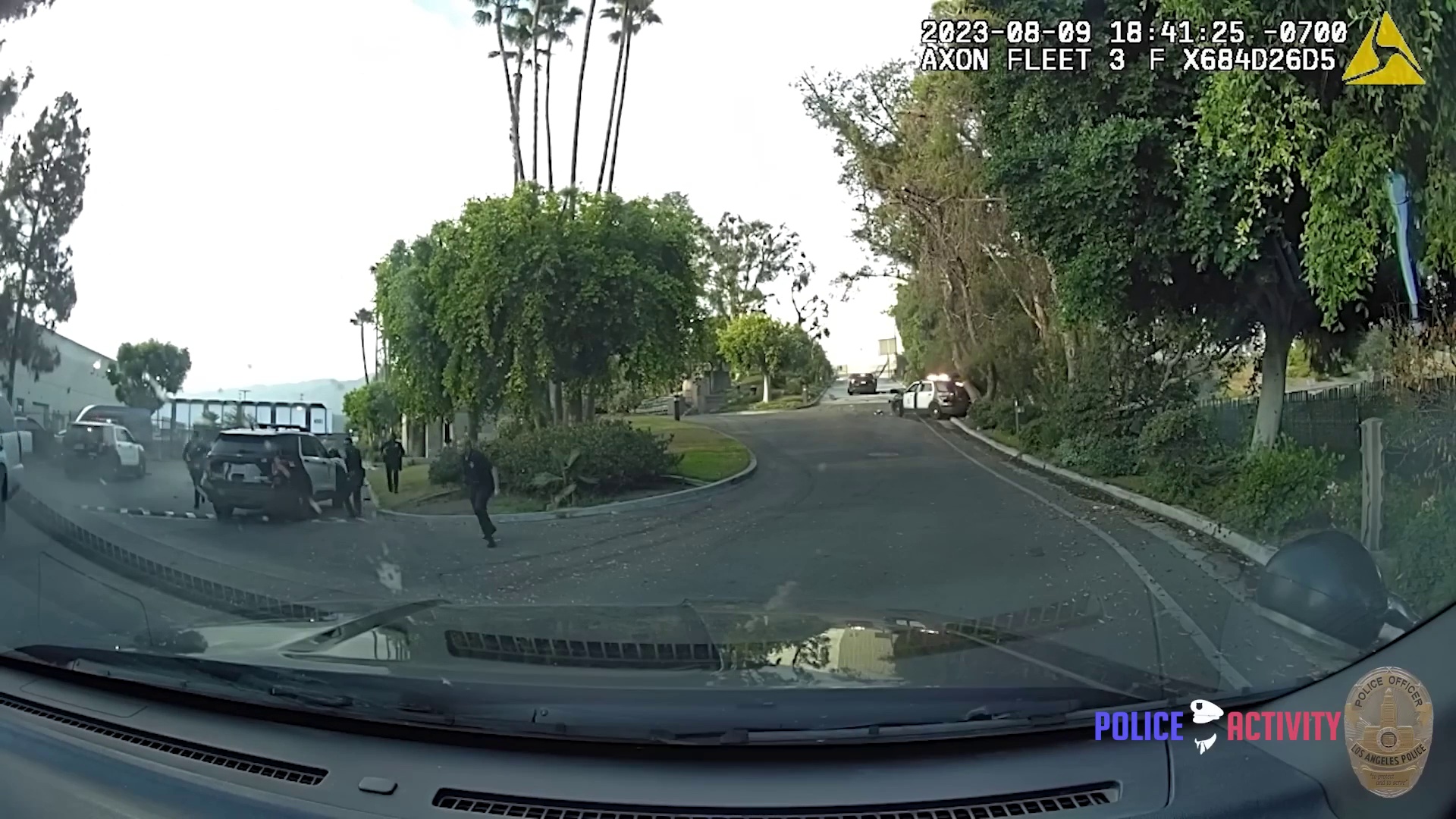}}
& Officers enter the scene; decisive contact is not visible.
& Repeated gunshots provide the decisive cue.
& NV & V \\
\addlinespace[1pt]
\raisebox{-0.48\height}{%
  \includegraphics[width=1.90cm]{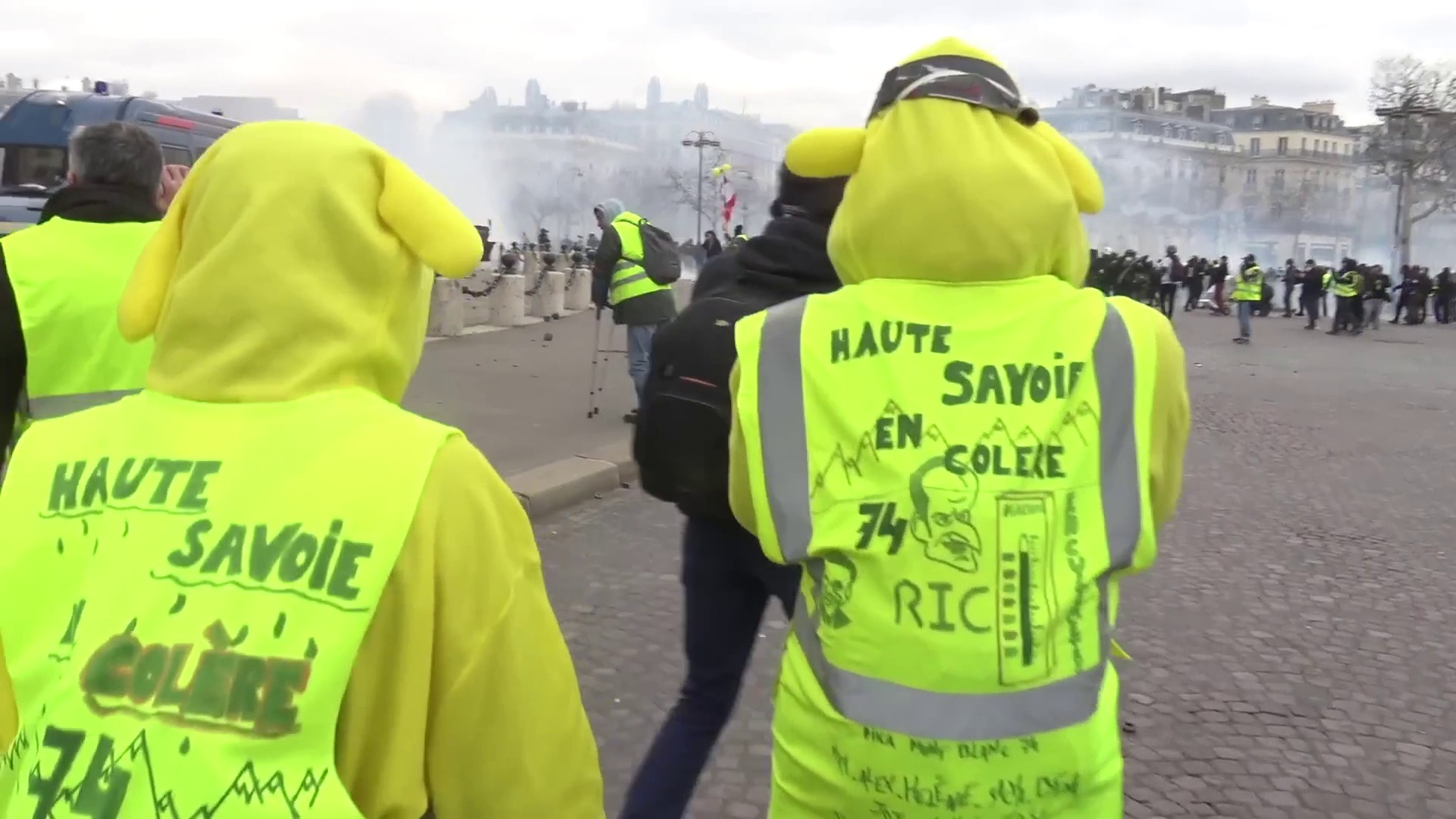}\hspace{1pt}%
  \includegraphics[width=1.90cm]{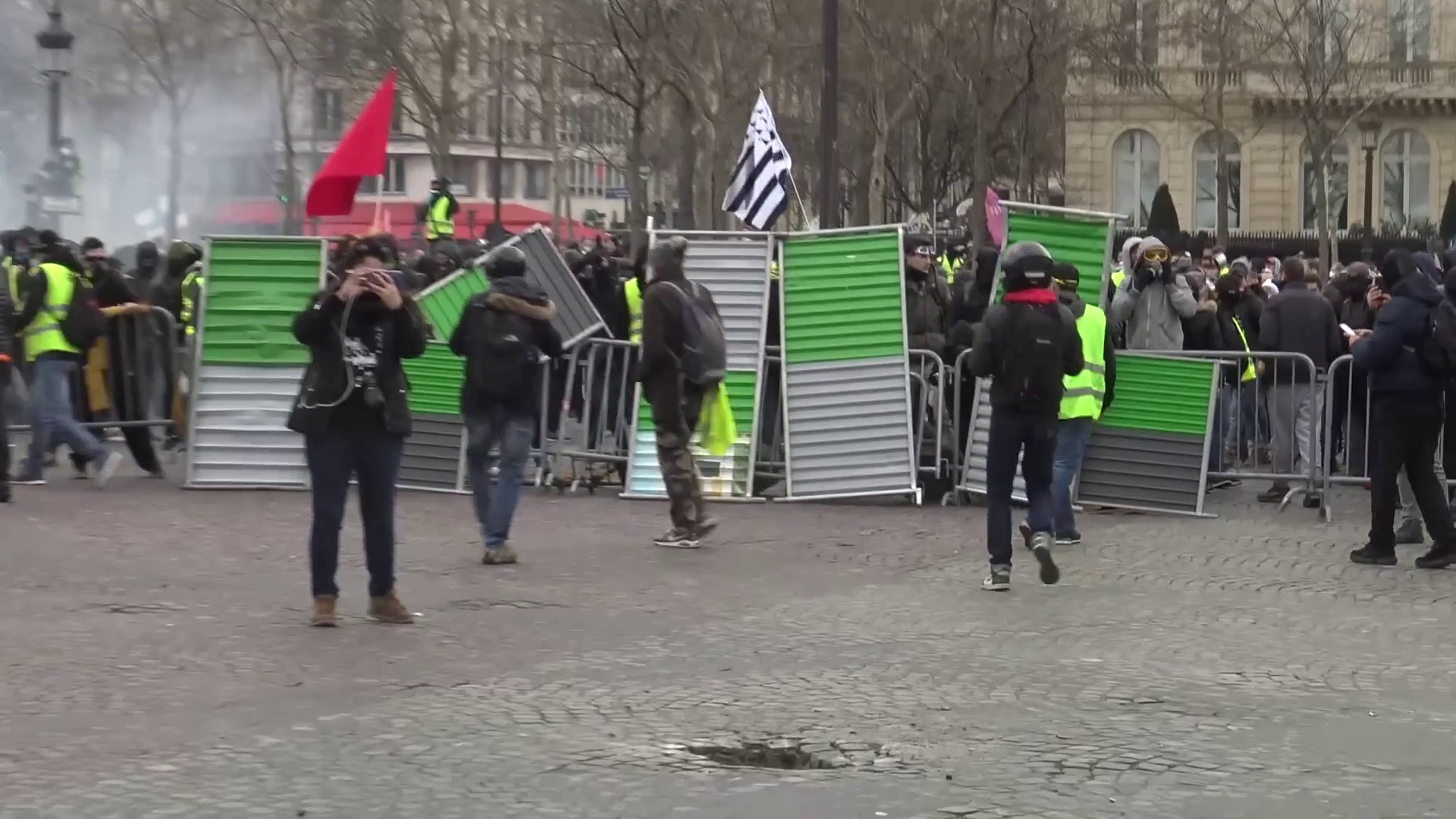}\hspace{1pt}%
  \includegraphics[width=1.90cm]{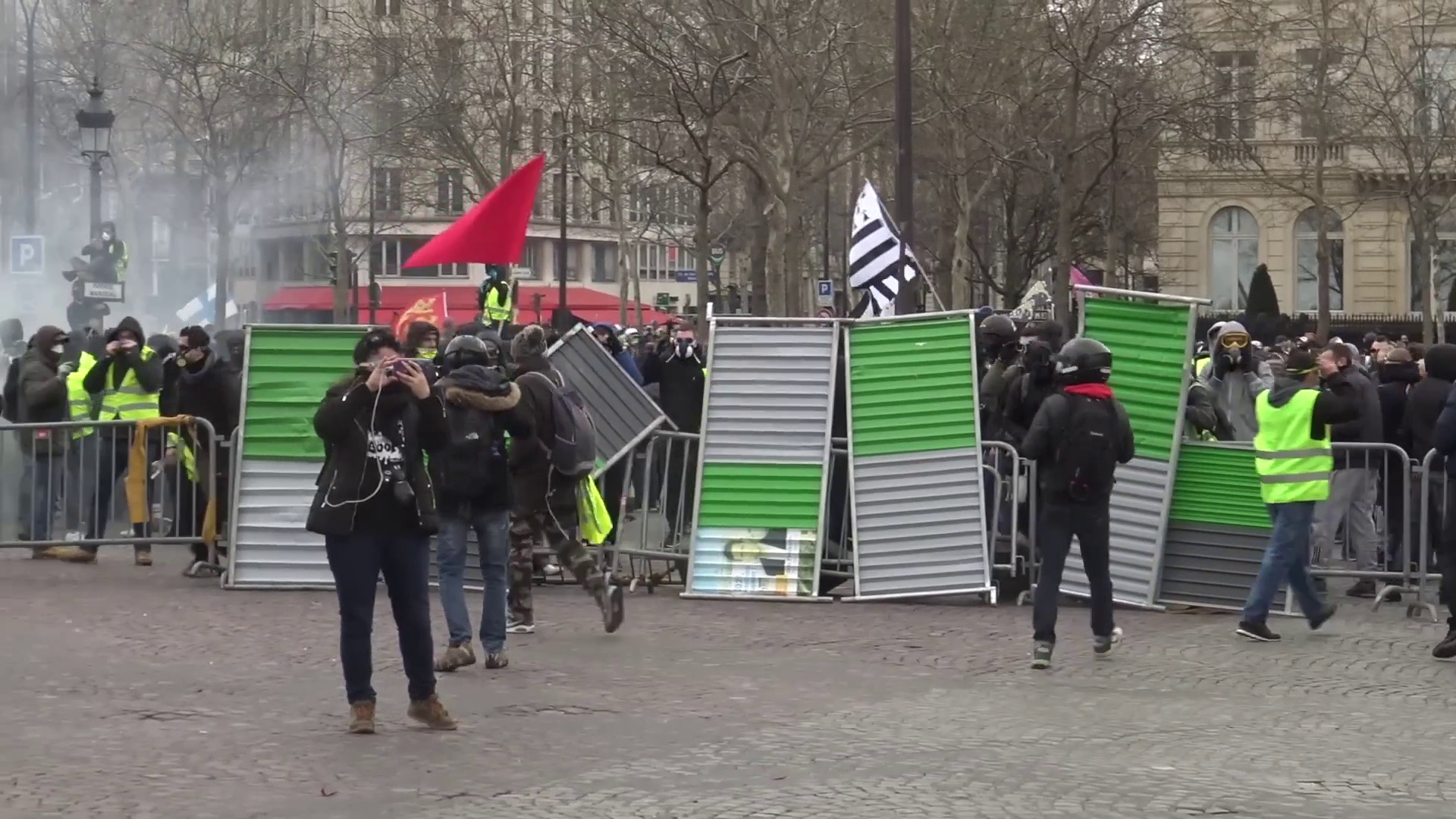}}
& Crowding and occlusion make the interaction ambiguous.
& Explosive sounds disambiguate the event.
& NV & V \\
\midrule
\multicolumn{5}{l}{\textit{Audio hurts}} \\
\raisebox{-0.48\height}{%
  \includegraphics[width=1.90cm]{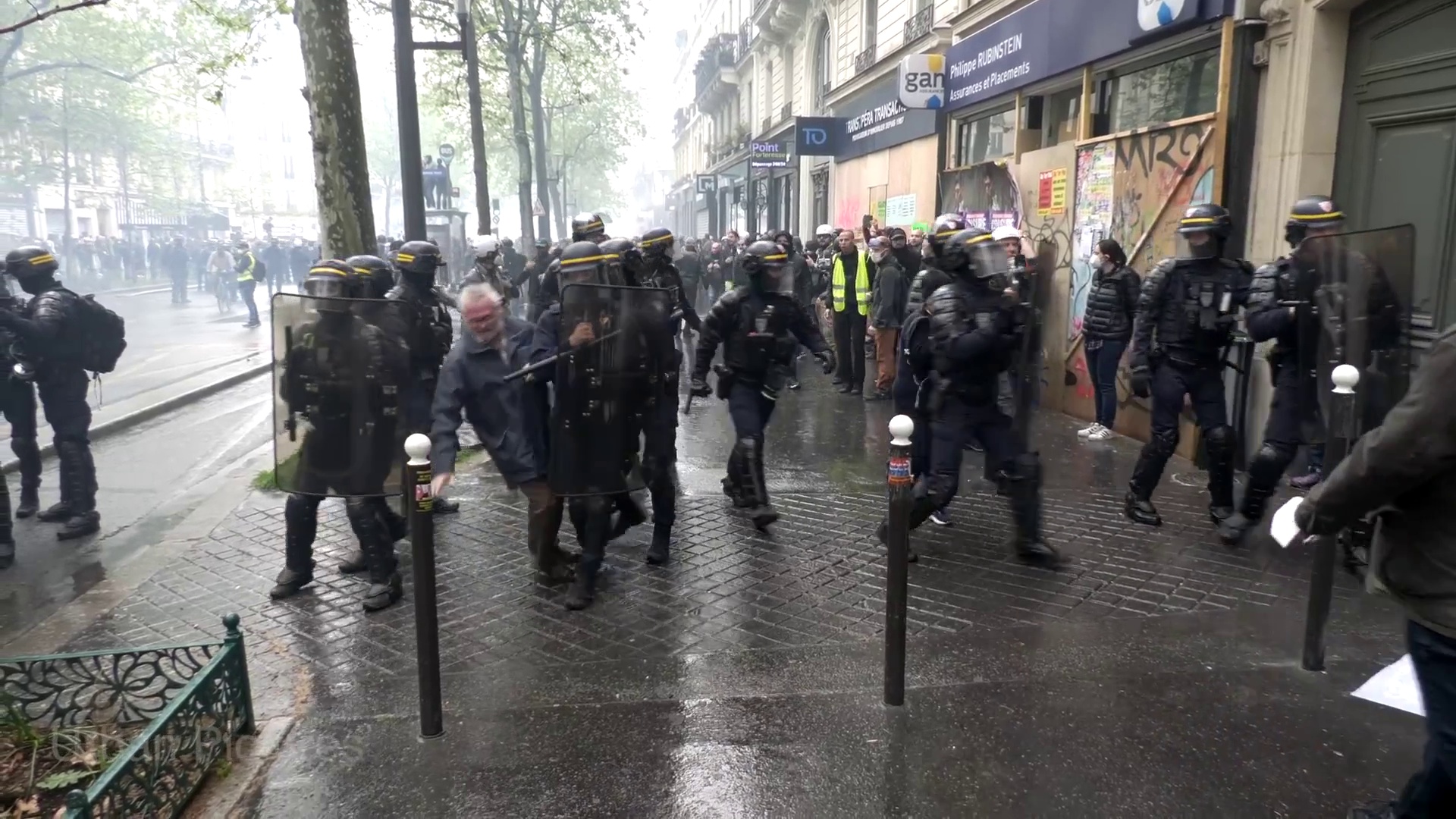}\hspace{1pt}%
  \includegraphics[width=1.90cm]{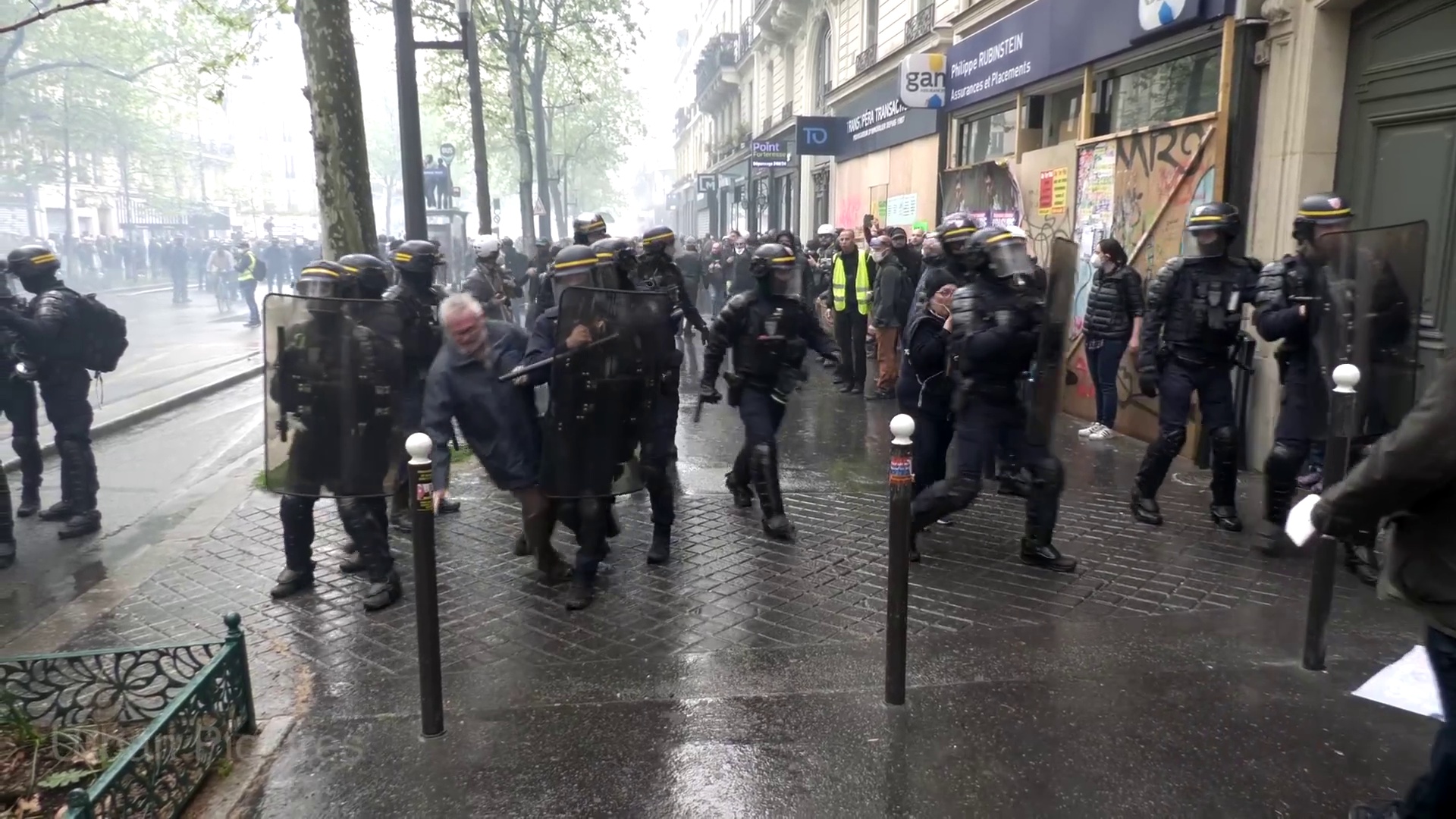}\hspace{1pt}%
  \includegraphics[width=1.90cm]{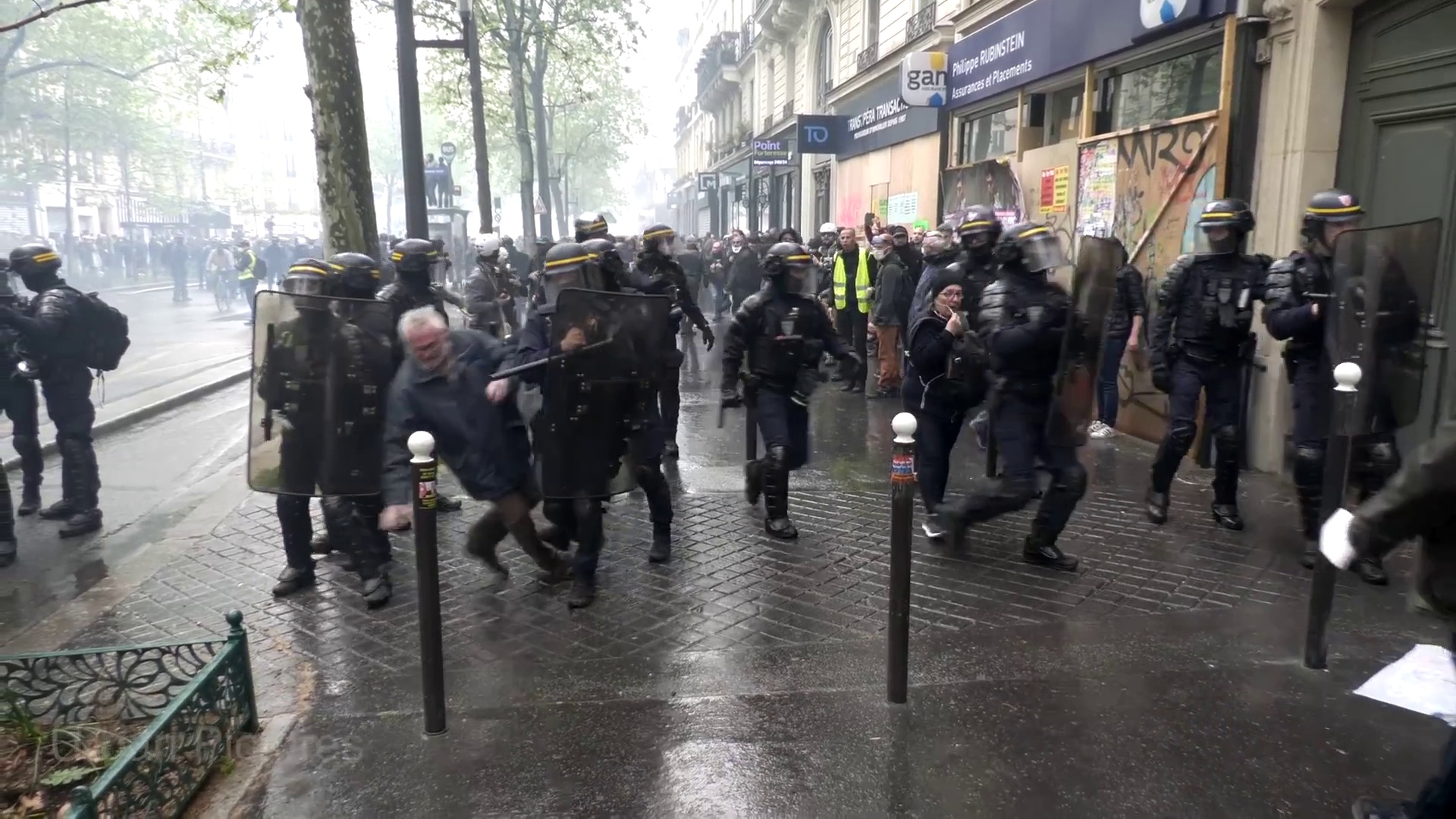}}
& A visible push causes a person to fall.
& Environmental noise masks event-related sound.
& V & NV \\
\addlinespace[1pt]
\raisebox{-0.48\height}{%
  \includegraphics[width=1.90cm]{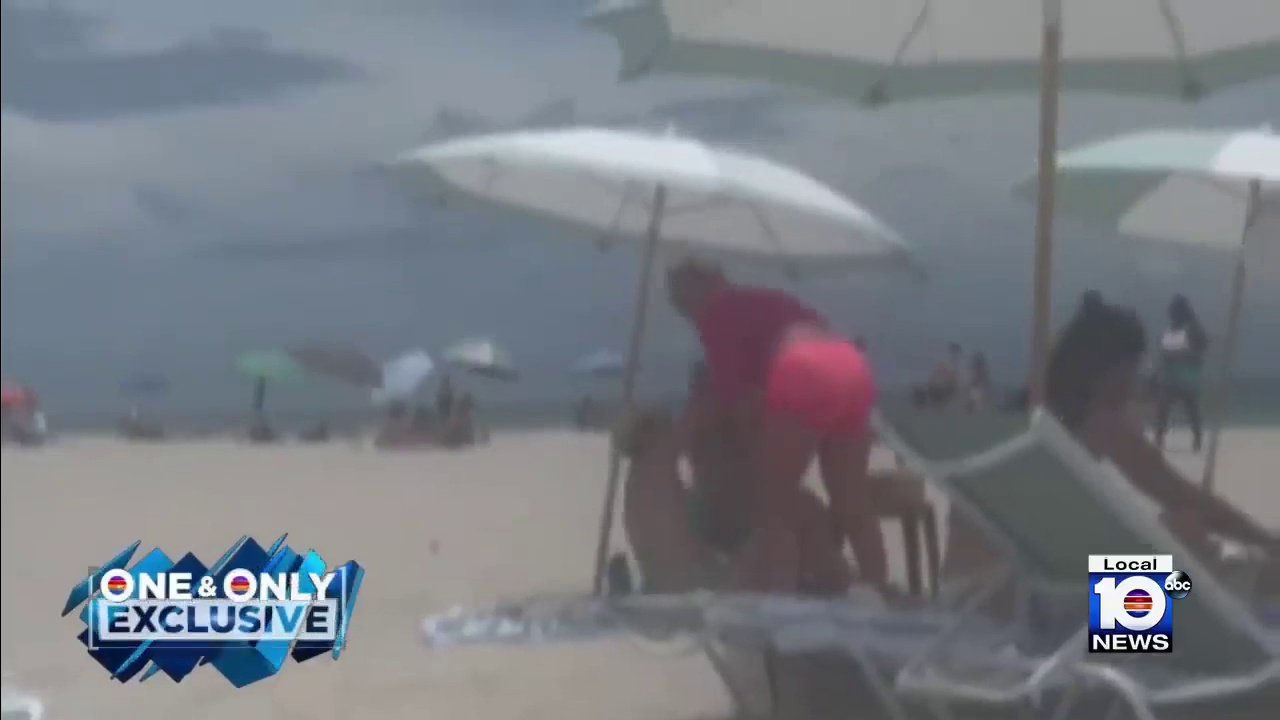}\hspace{1pt}%
  \includegraphics[width=1.90cm]{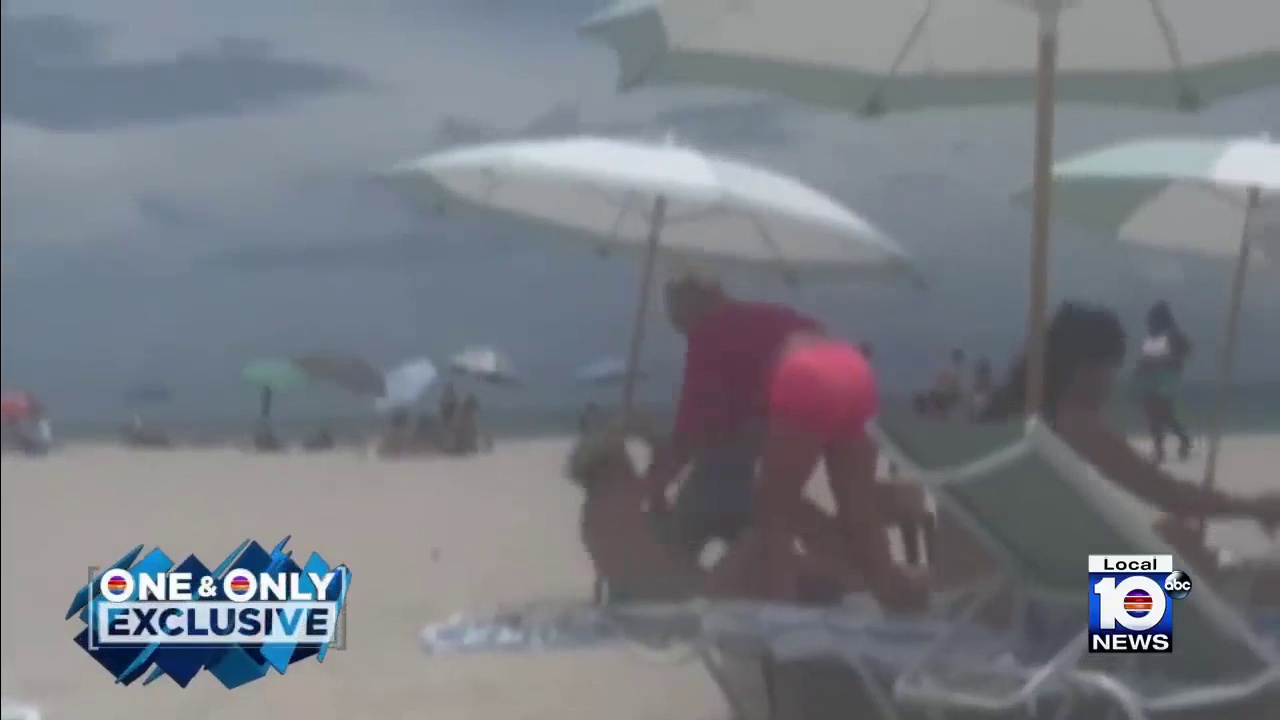}\hspace{1pt}%
  \includegraphics[width=1.90cm]{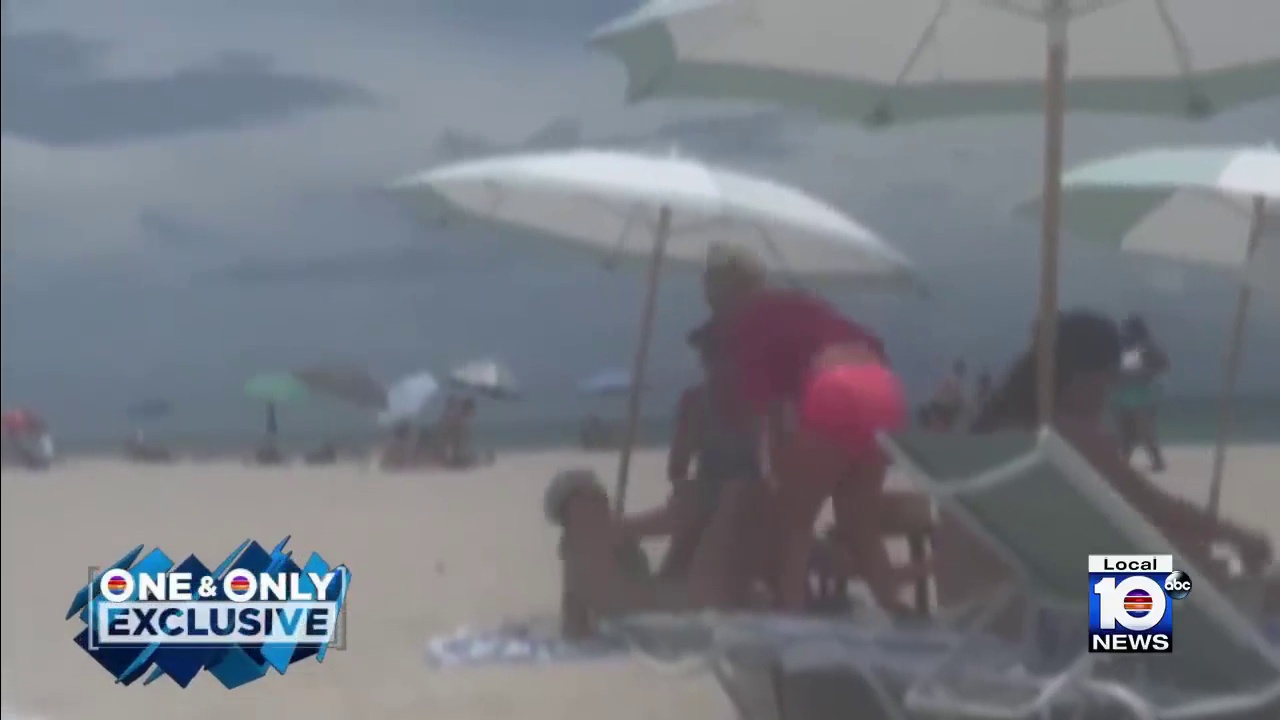}}
& Two people visibly strike one another.
& Wind and intermittent muting dominate the audio.
& V & NV \\
\bottomrule
\end{tabular}
\end{table*}

\subsection{Controlled Routing Intervention}

The ablations show that routing improves performance, but they do not separate the benefit of using audio from the benefit of learning when and how strongly to route it. We thus perform a controlled intervention on the same trained DVD checkpoint by replacing the learned route at inference time with fixed gate values. This keeps the model weights, test clips, and evaluation protocol unchanged, so the comparison isolates the effect of the routing policy. Table~\ref{tab:forced_gate} shows two clear patterns. First, fixed non-zero gates outperform the audio-disabled condition, confirming that the audio stream provides complementary evidence. Second, the learned route performs best, improving over both the audio-disabled and fixed-gate settings. The paired clip-level transitions further show that the learned route reduces errors overall rather than simply exchanging one error type for another.

Class-wise analysis provides additional insight into these changes. Compared with the audio-disabled condition, fixed routing reduces both violent misses and non-violent false alarms, showing that audio helps both classes. Learned routing improves this balance further: relative to fixed routing, it increases both violent and non-violent recall and reduces total errors. This indicates that the learned gate is not merely shifting the decision threshold toward one class, but improving the use of audio evidence across both classes.
This intervention therefore separates two effects: audio provides useful complementary information, and sample-dependent routing provides an additional gain beyond using audio with a constant weight.

\begin{table*}[h]
\centering
\caption{Matched forced-gate intervention on DVD. All metrics are percentages except the error count. The right panel reports paired prediction changes on the same test clips; C/I denotes errors corrected/introduced by the learned route.}
\label{tab:forced_gate}
\scriptsize
\begin{subtable}[t]{0.56\textwidth}
\centering
\caption{ Effect of fixed and learned routing.}
\setlength{\tabcolsep}{3.3pt}
\begin{tabular}{lrrrrr}
\toprule
Condition & Acc. & Bal. Acc. & Macro-F1 & ROC-AUC & Errors \\
\midrule
Video only & 71.99 & 71.36 & 71.47 & 79.20 & 163 \\
Forced $g=0$ & 72.51 & 71.90 & 72.01 & 79.90 & 160 \\
Forced $g=0.5$ & 74.74 & 74.21 & 74.32 & 82.10 & 147 \\
Forced $g=1$ & 74.57 & 73.98 & 74.11 & 81.80 & 148 \\
Learned $g$  & \textbf{75.74} & \textbf{75.43} & \textbf{75.44} & \textbf{83.30} & \textbf{141} \\
\bottomrule
\end{tabular}
\end{subtable}\hfill
\begin{subtable}[t]{0.41\textwidth}
\centering
\caption{  Paired prediction changes from forced gates to the learned gate.}
\setlength{\tabcolsep}{3pt}
\begin{tabular}{lrrrr}
\toprule
Comparison & Gain & C/I & Net clips & Errors after \\
\midrule
Learned vs $g=0$ & +3.78 & 35/13 & +22 & 141 \\
Learned vs $g=0.5$ & +1.55 & 22/13 & +9 & 141 \\
\bottomrule
\end{tabular}
\end{subtable}
\end{table*}

Table~\ref{tab:qualitative_flip} complements the aggregate results with representative prediction flips from the DVD analysis. The upper rows correspond to cases where the visual-only model fails but \method predicts correctly after incorporating audio, whereas the lower rows show instances where misleading audio overturns an otherwise correct visual prediction. These examples illustrate the two sides of audiovisual evidence. Audio helps when the visual stream is distant, crowded, or occluded and the soundtrack contains highly diagnostic cues such as gunshots or explosions. Conversely, strong environmental noise, wind, or intermittent muting can still degrade performance. The mechanistic analysis in the next subsection provides an explanation for this behavior: the learned routes redistribute visual-conditioned adaptation across layers rather than applying a single global accept-or-reject decision to the entire soundtrack.

\subsection{Mechanistic Analysis of Learned Routes}
\begin{figure*}[h]
    \centering
    \includegraphics[width=0.77\textwidth]{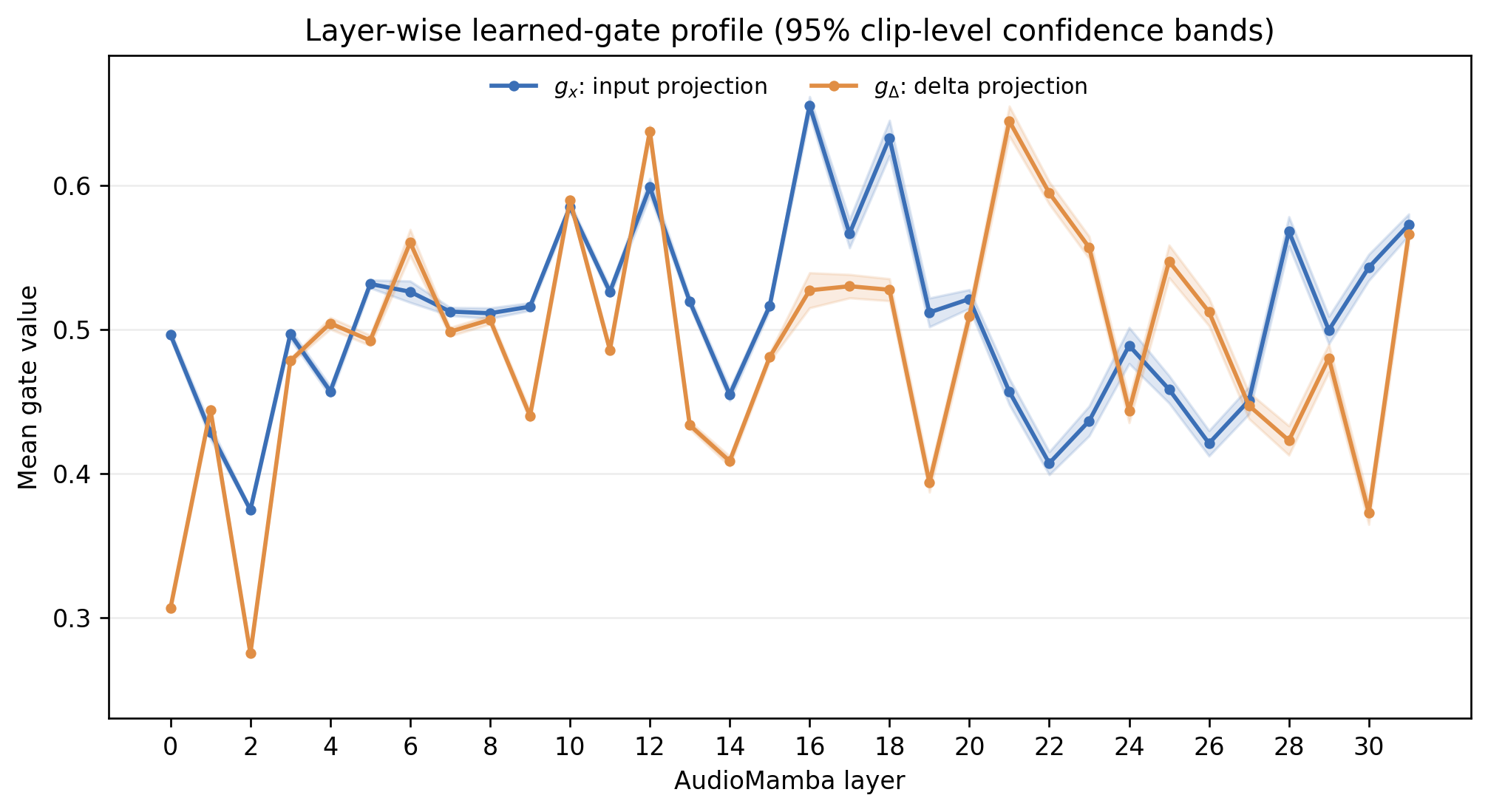}
    \caption{Mean post-sigmoid routing values across AudioMamba layers. Shaded bands show 95\% confidence intervals over clips after averaging the 12 evaluation views per clip. The input-parameter and step-size routes exhibit distinct depth profiles, indicating that they learn different layer-wise control patterns.}
    \label{fig:gate_layers}
\end{figure*}

The controlled intervention shows that the learned route improves performance over fixed routing policies. We next analyze the learned gates to understand what kind of routing behavior the model has acquired. Specifically, we ask whether the route acts as a simple global audio-quality switch, or whether it provides layer-specific control over the audio state-space operators. Figure~\ref{fig:gate_layers} visualizes the mean post-sigmoid routing value at each AudioMamba layer for the two modulated pathways: the input-parameter generator and the step-size generator. The learned routes are strongly depth-dependent, and the two pathways follow distinct layer-wise profiles. This indicates that they learn different control functions rather than sharing a single routing pattern.

Table~\ref{tab:gate_stats} summarizes the same routing values statistically. It reports their mean, standard deviation, range, saturation count and a variance decomposition into layer, clip and clip-by-layer effects. The gates are rarely saturated, with almost all values lying inside the open interval $[0.05,0.95]$, showing that the model does not simply learn to always accept or always suppress audio. The variance decomposition provides the main mechanistic insight. Most of the routing variation is explained by layer structure and clip-by-layer interaction, while the global clip effect is small. This means that the route is not merely estimating whether a clip has `good' or `bad' audio. Instead, it decides where in the audio hierarchy visual conditioning should be applied for each clip.

\begin{table}[h]
\centering
\caption{
Routing activation and variance decomposition. ``Sat.'' reports the number of routing values outside $[0.05,0.95]$. Layer, Clip and C$\times$L denote the percentage of variance explained by layer, clip and clip--layer interaction effects, respectively.
}
\label{tab:gate_stats}
\small
\setlength{\tabcolsep}{3pt}
\resizebox{\columnwidth}{!}{%
\begin{tabular}{lrrrrrrr}
\toprule
Path & Mean & SD & Range & Sat. & Layer & Clip & C$\times$L \\
\midrule
Input gen. & 0.507 & 0.153 & 0.060--0.953 & 1 & 33.5 & 1.4 & 65.1 \\
Step-size gen. & 0.488 & 0.150 & 0.047--0.941 & 4 & 48.7 & 1.2 & 50.0 \\
\bottomrule
\end{tabular}}
\end{table}

\subsection{Robustness to Invalid Audio \& Cross-Dataset Transfer}
The audio-valid benchmark evaluates fusion when both streams are present and scene-related. We next test two more challenging settings: unfiltered audio at test time and zero-shot transfer across datasets. These experiments assess whether \method remains effective when the audio stream may be missing, silent, or off-scene, and whether the learned representation transfers beyond the training dataset.

Table~\ref{tab:robust_native} reports results on the full unfiltered DVD test split. This evaluation is intended as a native-inference robustness study rather than a complete benchmark comparison. The goal is to test whether \method remains stable when the audio stream is missing, silent, or not scene-related, while keeping the controlled audio-valid benchmark as the primary comparison against SOTA A/V methods.\footnote{We use SlowFast-AV as is the only competing multimodal method whose native inference procedure can process such clips without introducing an additional missing-modality policy. Transformer-based methods in Table~\ref{tab:main_results} assume both modalities are present; handling invalid audio would require choices (e.g., masking, zero-imputation), which would alter their native inference behavior and create new method variants, and confound robustness with implementation-specific missing-modality handling.}

\begin{table}[h]
\centering
\caption{Native unfiltered-audio robustness on DVD (279 V / 382 NV).}
\label{tab:robust_native}
\begin{tabular}{lrrrr}
\toprule
Model & Acc. & F1-V & F1-NV & Macro \\
\midrule
SlowFast-AV & 63.54 & 56.10 & 68.82 & 62.46 \\
Route off + balanced & 73.22 & 63.51 & 78.85 & 71.18 \\
Route on + balanced & 74.58 & 64.71 & 80.14 & 72.42 \\
Route on + adaptive & \textbf{75.19} & \textbf{65.25} & \textbf{80.71} & \textbf{72.98} \\
\bottomrule
\end{tabular}
\end{table}

\method achieves the best performance among the evaluated variants. Routing and adaptive alignment improve performance monotonically, suggesting that the proposed components help suppress misleading audio evidence without discarding useful acoustic cues.

Table~\ref{tab:generalization_zero_shot} reports zero-shot cross-dataset transfer. The strongest transfer direction is from DVD to NTU-CCTV. We attribute this to the greater visual and acoustic diversity of DVD, which contains real-world footage, media content and a broader range of environments, allowing the model to learn more transferable audio--visual representations. In contrast, NTU-CCTV is more homogeneous and thus transfers less effectively to the more diverse DVD setting.
Overall, results suggest that \method learns transferable audio--visual representations, while also highlighting the strong influence of dataset-specific characteristics, annotation protocols and supervision settings on cross-domain violence detection.

\begin{table}[h]
\centering
\caption{Zero-shot cross-dataset generalization.}
\label{tab:generalization_zero_shot}
\begin{tabular}{llrr}
\toprule
Train & Test & Acc. & Macro \\
\midrule
NTU & DVD & 61.17 & 52.72 \\
DVD & NTU & \textbf{69.59} & \textbf{69.56} \\
\bottomrule
\end{tabular}
\end{table}

\section{Conclusion}
We presented \method, a multimodal violence detection framework that uses visual context to adaptively modulate audio state-space operators. Across NTU-CCTV and DVD, \method consistently outperformed SOTA while requiring fewer parameters and lower computational cost than the leading Transformer-based methods. Controlled ablations and routing interventions showed that the gains arise not only from incorporating audio, but also from learning when and how strongly audio information should influence the model. Mechanistic analyses further revealed that the learned routes act as layer-specific controllers rather than as a single global audio weighting mechanism. Robustness experiments demonstrated that adaptive routing remains beneficial under imperfect audio conditions and enables meaningful transfer across datasets. Future work will explore extending operator-level conditioning to other A/V tasks and to settings with more severe modality corruption or missing inputs.

{\small
\bibliographystyle{ieeenat_fullname}
\bibliography{references}
}

\end{document}